\documentclass[conference]{IEEEtran}
\IEEEoverridecommandlockouts

\usepackage{cite}
\usepackage{amsmath,amssymb,amsfonts}
\usepackage{booktabs}
\usepackage{graphicx}
\usepackage{xcolor}
\usepackage{url}
\usepackage{array}
\usepackage{multirow}

\newcolumntype{C}[1]{>{\centering\arraybackslash}p{#1}}

\newcommand{\zraw}{z_{\mathrm{raw}}}
\newcommand{\zlrg}{z_{\mathrm{LRG}}}
\newcommand{\rfull}{r_{\mathrm{full}}}
\newcommand{\rlda}{r_{\mathrm{lda}}}
\newcommand{\wtlp}[2]{#1~(#2)}

\begin{document}

\title{Local Reference Geometry Residual Augmentation for Imbalanced Time Series Classification}

% \author{
% \IEEEauthorblockN{Anonymous Authors}
% \IEEEauthorblockA{Anonymous Institution\\
% Email hidden for review}
% }
\author{
\IEEEauthorblockN{
Chuanhang Qiu,
Yanran Xu,
Yue Wang,
Anthony Bagnall
}
\IEEEauthorblockA{
School of Electronics and Computer Science\\
University of Southampton\\
Southampton, United Kingdom\\
Emails: C.Qiu@soton.ac.uk, Y.Xu@soton.ac.uk,
Yue.Wang@soton.ac.uk, A.J.Bagnall@soton.ac.uk
}
}
\maketitle

\begin{abstract}
Imbalanced time series classification is often addressed by changing the training distribution, objective, logits, or final threshold. These interventions address important biases, yet leave a representation-level question unmeasured: after minority support is reduced, does a learned feature space remain locally reliable around minority regions? We identify a training-local geometry failure: under imbalance, minority cases can lie in sparse, rest-dominated, or mixed feature-space neighborhoods, even when the representation retains useful global class structure. To diagnose and repair this failure, we propose Local Reference Geometry (LRG), a lightweight post-hoc feature augmentation module applied between a fixed feature extractor and the classifier head. Using training features only, LRG measures local exposure and class-mixture risk, then augments each fixed feature with a standardized signed displacement from nearby training geometry and an LDA-projected residual summary. On controlled UCR/Bake Off Redux imbalance benchmarks, paired raw-versus-LRG comparisons show gains for learned, pretrained, and fixed representations, including when LRG is combined with training-level interventions and post-encoder classifier corrections. Ablations show that the gain comes from the signed local residual appended to the original feature, rather than from generic prototype distances, affinity features, scalar statistics, or VLAD-style codes. Further analyses support the proposed local-geometry failure hypothesis: minority neighborhoods become increasingly rest-exposed under imbalance, training-local risk identifies error-prone regions, and LRG gains concentrate in those high-risk regions. 
%Code and reproducibility files are provided at \url{https://anonymous.4open.science/r/LRG}.

\end{abstract}

\begin{IEEEkeywords}
imbalanced time series classification, representation learning
\end{IEEEkeywords}

\section{Introduction}

Many labeled time series datasets include rare classes that correspond to practically important events, such as equipment faults, medical abnormalities, or other safety-critical anomalies. The resulting class imbalance limits minority support during training and can make rare-event recognition unreliable. Existing approaches address this problem at different stages of the pipeline. Data-level methods alter the training distribution, but require access to the input series and rely on assumptions about temporal similarity \cite{zhu2022minority}. Training-level methods change the sampler, objective, or training procedure, and usually require retraining the encoder \cite{geng2018costsensitive}. Post-encoder methods adjust the final decision rule while leaving the feature space fixed \cite{menon2021logitadjustment}. These interventions address important parts of the imbalance problem, but they do not test whether the learned representation remains locally reliable for recognizing minority samples when training support is scarce.

Long-tailed recognition~\cite{zhang2023deeplongtail} separates classifier-level correction from representation-level effects in computer vision: decoupling studies show that classifier correction can be effective after a representation has been learned~\cite{kang2020decoupling}, while representation-learning studies show that imbalance can also distort tail-class feature geometry~\cite{zhu2022bcl}. This raises the corresponding question for imbalanced TSC: given an existing time-series representation, can we identify and correct local minority unreliability without retraining the encoder?

In this post-hoc setting, the relevant unreliability is local to the learned representation. A minority sample may lie in a region where nearby training support is sparse, mixed across classes, or dominated by non-minority samples, even when the feature space still separates classes at a coarser scale. A reweighted or calibrated classifier can adjust priors or thresholds, but it still operates on the features already available from the representation. We study this missing local information as a training-local geometry failure. It can be measured from the training set alone by comparing minority exposure with the training class distribution and by estimating class mixing in local regions.

To address this training-local failure after feature extraction, we propose Local Reference Geometry (LRG), a residual augmentation method for fixed representations. LRG is a supervised post-hoc feature augmentation method that uses training features and labels while leaving the encoder or transform unchanged. It builds local reference regions in the training representation, then augments each feature vector with a signed local residual and a compact supervised residual summary. These added features provide information about local training support that may not be explicit in the original representation. The same reference regions also provide diagnostics of local exposure and class-mixture risk. LRG is therefore useful when features have already been produced by a learned encoder, a pretrained model, or a fixed transform, and retraining the feature extractor is costly or impractical.

Our contributions are:
\begin{itemize}
\item We define and measure training-local geometry failure for imbalanced
TSC using minority-neighborhood exposure and training-local
class-mixture risk.
\item We propose LRG, a supervised post-hoc residual augmentation method
fitted from training features and labels. LRG augments each fixed feature
vector with a signed local residual and an LDA-projected residual summary,
while leaving the encoder or transform unchanged.
\item We evaluate LRG on controlled UCR/Bake Off Redux multi-ratio tasks,
showing paired gains across learned, pretrained, and fixed representations,
compatibility with imbalance interventions and post-encoder corrections,
and larger gains in regions identified as high risk by training-only
diagnostics.
\end{itemize}
Code and reproducibility files are provided in the associated repository\footnote{\url{https://anonymous.4open.science/r/LRG}}.

% Fig.~\ref{fig:lrg-landscape-intro} summarizes this positioning relative to
% data-level, training-level, and post-encoder decision remedies.
% \begin{figure*}[!t]
% \centering
% \includegraphics[width=\textwidth]{figures/lrg_landscape_intro.pdf}
% \caption{LRG in the imbalanced TSC remedy landscape. Data-level and
% training-level remedies act before or during representation learning, while
% post-encoder decision remedies adjust the downstream decision rule after features are
% fixed. LRG targets a complementary feature-space failure mode: it builds
% training-only local reference regions, measures exposure and class mixing, and
% augments the frozen feature with signed residual coordinates.Need redrawn!}
% \label{fig:lrg-landscape-intro}
% \end{figure*}

\section{Related Work}

Related work falls into three groups: data-level and training-level remedies for
imbalanced TSC, post-encoder decision rules and long-tailed correction, and
feature-space geometry or local diagnostic representations. LRG is applied
after feature extraction and is complementary to these intervention points.

\subsection{Imbalanced Time Series Remedies}

Imbalanced time series classification methods are often described using a
coarser data-level versus algorithm-level taxonomy
\cite{zhu2022minority,geng2018costsensitive}. Under the intervention-point view
used here, data-level remedies change the training set through under-sampling,
over-sampling, or synthesis, including
SMOTE \cite{chawla2002smote}, time-series variants such as e-SMOTE~\cite{qiu2025esmote} and T-SMOTE 
\cite{zhao2022tsmote}. Training-level remedies change the sampler, learner,
objective, or training-time logits, including focal loss \cite{lin2017focal}, Balanced Softmax
\cite{ren2020balancedsoftmax}, LDAM \cite{cao2019ldam}, and supervised or
balanced contrastive objectives \cite{zhu2022bcl}. These
baselines act before or during encoder/classifier training. LRG instead asks
whether an already produced feature space contains minority-relevant local
failures that can be exposed to a downstream head without retraining the
encoder.

\subsection{Long-Tailed and Post-Encoder Correction}

Long-tailed recognition in computer vision separates two effects of class
imbalance. Decoupling and calibration studies show that, once a representation
is fixed, correcting the classifier head can be a strong intervention; balanced
representation learning studies show that imbalance can also shape tail-class
feature geometry
\cite{zhang2023deeplongtail,kang2020decoupling,zhu2022bcl}.
This distinction is useful for LRG: classifier correction is a strong
frozen-feature baseline, and the relevant question is whether an additional
local residual coordinate still provides information after the decision rule
has been corrected. Recent time series work has begun to develop related
representation-level objectives \cite{jin2023timecontrastive,qian2025dgmscl}.
LRG is aligned with this representation-level view, but it is post-hoc: the
feature extractor is held fixed and the correction is expressed as an added
feature coordinate.

Post-encoder methods also operate after features have been extracted, but they
act on the classifier head or decision rule. Logit adjustment \cite{menon2021logitadjustment},
Prior2Posterior \cite{bhat2025prior2posterior}, temperature or vector scaling
\cite{guo2017calibration}, classifier re-training, and classifier weight
normalization \cite{kang2020decoupling} correct priors, logits, calibration, or
classifier weights. Under a frozen-feature constraint, these methods are close
baselines and we evaluate them in Table~\ref{tab:post-encoder}. Their role is
different from LRG: they change how existing coordinates are read, whereas LRG
exposes a coordinate describing how a sample is displaced relative to nearby
training support. This is why our post-encoder study uses paired comparisons
within each correction rule, testing whether LRG remains useful when a
classifier-level correction is already applied.

\subsection{Feature-Space Geometry and Diagnostic Representations}

At a computational level, LRG is close to methods that introduce anchors,
prototypes, or projections in feature space. Prototype, metric-learning, and
discriminant methods learn class prototypes, metrics, or projections for
classification
\cite{snell2017prototypical,weinberger2009lmnn,sugiyama2007lfda}.
Radial-basis features, local coordinate coding, and VLAD-style residual
encodings similarly represent inputs through affinities, local codes, or
aggregated residuals from a codebook
\cite{broomhead1988multivariable,wang2010locality,jegou2010aggregating}. These
methods are designed as general classifiers, projections, or feature encodings.
LRG differs in problem setting: it does not learn a prototype classifier, a
metric, or a new representation. Instead, it asks whether local reference
geometry can expose minority-relevant residual information after a
representation has already been learned. This motivates the local-feature
controls against prototype distances, $k$-means distances, RBF affinities, scalar
local statistics, and a VLAD-style encoding.

Local-neighborhood diagnostics characterize class overlap, instance hardness,
and minority difficulty from neighborhood composition, including
data-complexity measures, $k$-disagreeing-neighbor statistics, and ADASYN
\cite{ho2002complexity,smith2014instance,he2008adasyn}. LRG shares this
training-neighborhood view but uses it for a different purpose: relative rest
exposure normalizes local composition by the training prior, and the signed
residual uses the same local frame to produce a repair coordinate. Thus, LRG is
best viewed as post-hoc feature-space repair: it is computationally related to
anchor-based feature encodings, but it targets the training-local reliability of a
fixed representation and complements sampling, loss-level, and classifier-level
corrections.

\section{Problem Setup}

\subsection{Task Definition and Evaluation Metrics}

We consider an imbalanced time series task with labeled training split
$\mathcal{D}_{\mathrm{tr}}=\{(x_i,y_i)\}_{i=1}^n$ and evaluation split
$\mathcal{D}_{\mathrm{te}}$. A representation extractor maps each time series
to a feature vector $z=f(x)\in\mathbb{R}^d$. The extractor may be a fixed
transform or a learned encoder. The classifier head is trained on training
features and evaluated on features from $\mathcal{D}_{\mathrm{te}}$.
For each dataset-ratio task, the minority class is defined from the training
split:
\begin{equation}
    y_{\min} = \arg\min_c |\{i:y_i=c\}|.
\end{equation}
Our benchmark keeps the original label space. In a multi-class source dataset,
only $y_{\min}$ is treated as the minority class; the remaining classes are kept
as separate classes during head training rather than being collapsed into a
single negative class. Minority-oriented metrics are then computed by
binarizing predictions into $y_{\min}$ versus rest. Sens. denotes minority
recall, Spec. denotes specificity, and Min-F1 is the F1 score of $y_{\min}$.

\subsection{Training-Local Geometry Diagnostics}

To quantify training-local reliability, we measure how often minority samples
are surrounded by non-minority examples in the learned feature space. For each
minority training sample $i$, let $q_i$ be the fraction of its $k=5$ nearest
other training samples that do not belong to $y_{\min}$. The raw local
rest-neighbor fraction is $\mathbb{E}_{i:y_i=y_{\min}}[q_i]$. This raw
fraction depends on the rest-class frequency of the specific imbalanced split.
We therefore normalize it by
$p_{\mathrm{rest}}=n_{\mathrm{rest}}/(n_{\mathrm{tr}}-1)$, the
random-neighbor expectation for a minority query under the same training split,
where $n_{\mathrm{rest}}$ is the number of non-minority training samples, and
define relative rest exposure (RRE) as
\begin{equation}
    \mathrm{RRE} =
    \frac{\mathbb{E}_{i:y_i=y_{\min}}[q_i]}{p_{\mathrm{rest}}}.
    \label{eq:rre}
\end{equation}
RRE equals one when the local rest-neighbor rate matches the split-specific
rest-class prior; larger values indicate that minority neighborhoods are more
rest-exposed than expected from the prior alone. For a collection of
dataset-ratio tasks, RRE is computed within each task first and then averaged
across tasks; it is not the ratio of an averaged local fraction to an averaged
global prior. All neighborhood diagnostics are computed only from training
features and labels.

\section{Local Reference Geometry}

LRG uses one training-local reference frame for both diagnosis and repair.
Scalar outputs, including exposure and class-mixture risk, diagnose locally
unreliable regions; vector outputs provide signed residual coordinates for
repair. This section defines the repair feature, while the analysis section uses
the same frame to test whether risky regions coincide with raw prediction
failures.

\subsection{Local Reference Regions}

Let $f$ denote a fixed feature extractor; learned encoders are trained before
LRG is fit and are not updated by LRG. LRG operates on flattened features that
are standardized with training-split statistics and then $\ell_2$-normalized.
Let $Z_{\mathrm{tr}}=\{z_i=f(x_i):(x_i,y_i)\in\mathcal{D}_{\mathrm{tr}}\}$ be
the resulting training features. LRG partitions $Z_{\mathrm{tr}}$ into
$K_{\mathrm{eff}}$ $k$-means regions \cite{lloyd1982least}:
\begin{equation}
    \{c_1,\ldots,c_{K_{\mathrm{eff}}}\}
    = k{\text{-means}}(Z_{\mathrm{tr}}, K_{\mathrm{eff}}).
\end{equation}
We use $K_{\mathrm{eff}}=\min(K,\max(2,\mathrm{round}(\sqrt{n_{\mathrm{tr}}})),
n_{\mathrm{tr}})$ with default cap $K=8$. Each region stores its center, a
radius $\rho_k$ given by the median assigned-sample distance with floor
$\epsilon=10^{-3}$, and its assigned training members. Labels are used only for
the LDA residual and diagnostic class-composition analyses, not for the $k$-means
regions. 

$k$-means serves as a stable, training-only anchoring mechanism rather than a class
model. Its centers are not minority prototypes, and LRG does not require
minority-dominated anchors. A rest-dominated region can itself expose the
failure mode: the signed residual describes how a sample departs from the nearby
training frame, while the downstream head learns whether that displacement is
discriminative. We avoid local covariance models or label-aware regions because
minority support is scarce under imbalance and high-dimensional local covariance
estimates can be unstable. The default frame therefore stores only centers and
scalar radii, avoids test-distribution information, and reserves labels for the
LDA projection and diagnostic risk analysis.

\subsection{Local Affinity and Residual}

For any feature $z$, define its soft affinity to local reference regions with
temperature $\tau>0$:
\begin{equation}
    a_k(z)=
    \frac{\exp(-\|z-c_k\|_2^2/\tau)}
         {\sum_j \exp(-\|z-c_j\|_2^2/\tau)}.
\end{equation}
The local reference center and scale are
\begin{equation}
    \mu(z)=\sum_{k=1}^{K_{\mathrm{eff}}} a_k(z)c_k,\quad
    \sigma(z)=\sum_{k=1}^{K_{\mathrm{eff}}} a_k(z)\rho_k.
\end{equation}
The local residual is
\begin{equation}
    \rfull(z)=\frac{z-\mu(z)}{\max(\sigma(z),\epsilon)}.
\end{equation}
This residual is a signed local-reference coordinate, not a prototype decision
rule. Soft affinities smooth hard cluster boundaries and let interface samples
inherit a reference center from several nearby anchors. The radial scale makes
residual magnitudes comparable across dense and diffuse regions, while the sign
and direction describe how the feature departs from the local training frame.

\subsection{Geometric Effect of Residual Augmentation}
\label{sec:geometric-effect}

The role of the residual can be seen by comparing the decision functions
available to a linear downstream head. A linear score on the raw feature has the
form
\begin{equation}
    s_y^{\mathrm{raw}}(z)=w_y^\top z+b_y ,
\end{equation}
so pairwise class decisions have global affine boundaries in the original
feature space. With residual augmentation, the score becomes
\begin{equation}
    s_y^{\mathrm{LRG}}(z)=w_y^\top z+v_y^\top \rfull(z)+u_y^\top \rlda(z)+b_y .
\end{equation}
Setting $v_y=u_y=0$ recovers $s_y^{\mathrm{raw}}(z)$, so residual augmentation
does not remove any decision function available on $\zraw$. The additional
terms are not arbitrary nonlinear capacity; they are tied to local displacement
from the training reference frame.
For intuition, consider the hard-assignment limit in which the nearest region
$k$ has $a_k(z)\to 1$ and all other regions have $a_j(z)\to 0$ for $j\ne k$.
Then the soft local mean and scale reduce to $\mu(z)\approx c_k$ and
$\sigma(z)\approx \rho_k$, so $\rfull(z)\approx (z-c_k)/\rho_k$.
Ignoring the low-dimensional LDA term for the moment,
\begin{equation}
    s_y^{\mathrm{LRG}}(z)
    \approx
    \left(w_y+\frac{v_y}{\rho_k}\right)^\top z
    + b_y-\frac{v_y^\top c_k}{\rho_k}.
\end{equation}
Thus, within a local reference region, the same linear head can express a
region-conditioned affine correction; with soft affinities, this correction is
smooth rather than a hard set of independent classifiers. The LDA term has the
same affine form because it is an affine function of $\rfull(z)$, as shown
below. This distinguishes LRG from scalar local statistics: affinity, purity, or
reliability can mark a region as near, mixed, or sparse, but a signed residual
specifies the direction of deviation from the local training frame. The downstream head
remains lightweight: LRG shares one downstream head, rather than introducing a
neural head or region-specific classifiers. Under the paired protocol, gains
after adding LRG reflect information exposed by the residual frame rather than a
wholesale change in model class.

\subsection{LDA-Projected Residual}

The residual can be high-dimensional, and not every direction is class-relevant.
We therefore summarize residuals with shrinkage-regularized LDA directions
$W_{\mathrm{LDA}}$ fit on training residuals and labels. The projection is
applied to residuals rather than raw features: raw-feature LDA would mainly
recover global class directions that a linear head can already learn, whereas
residual-space LDA identifies class-aligned directions of local deviation. This
LDA fit is global over the training residuals, not separate within each $k$-means
reference region, so it does not require estimating class-conditional covariance
matrices inside sparse local clusters. With training-split mean
$m_{\mathrm{LDA}}$ and diagonal standard deviation matrix $D_{\mathrm{LDA}}$,
the projected residual is
\begin{equation}
\begin{aligned}
    p_{\mathrm{LDA}}(z)&=W_{\mathrm{LDA}}^\top\rfull(z),\\
    \rlda(z)&=D_{\mathrm{LDA}}^{-1}
    \left(p_{\mathrm{LDA}}(z)-m_{\mathrm{LDA}}\right).
\end{aligned}
\end{equation}
Since $\rlda(z)$ is affine in $\rfull(z)$, adding it preserves the local affine
interpretation above. For binary tasks this is one-dimensional; for a $C$-class
task it can contain at most $C-1$ directions. We use a shrinkage LDA solver
(``lsqr'' with automatic shrinkage). All standardization, LDA fitting,
projection normalization, and head fitting are training-only. In all main
experiments, we use fixed local geometry parameters: $K=8$, retained LDA
dimension $q=1$, and $\tau=1.0$.
Because $\rlda(z)$ is appended to $[\zraw,\rfull(z)]$, the one-dimensional
default acts as a low-variance supervised residual hint rather than a bottleneck
or independent nonlinear map.
\subsection{LRG Residual Augmentation and Complexity}

The final feature passed to the downstream head is
\begin{equation}
    \zlrg = [\zraw,\rfull(z),\rlda(z)].
\end{equation}
LRG then trains the same lightweight heads used for $\zraw$, mainly
class-weighted logistic regression and threshold-tuned logistic regression. The
LRG reference frame--the standardizer, centers, radii, and LDA projection--is
built using only training features and labels, and is then kept fixed. At
evaluation time, each feature is standardized with training statistics, assigned
soft affinities to the stored centers, converted into residual coordinates, and
passed to the fitted head. The procedure uses no test labels and estimates no
test-set prior, matching the information constraint of the post-encoder
baselines.

Let $n$ be the number of training samples, $d$ the feature dimension, $C$ the
number of classes, $I$ the $k$-means iteration count, and $q\le C-1$. Reference
construction costs $O(InK_{\mathrm{eff}}d)$, and the persistent LRG state stores
$O(K_{\mathrm{eff}}d+dq)$ values plus the feature standardizer and linear head.
At test time, each sample costs $O(K_{\mathrm{eff}}d+dq)$ for the local frame
and LDA residual, plus $O(C(2d+q))$ for the linear head. Once features are
extracted, this overhead is independent of the original series length: LRG
stores no raw series, pairwise training distances, or nearest-neighbor graphs at
inference time.
\section{Experiments}

The experiments evaluate LRG through paired raw-versus-LRG comparisons and
compatibility checks with imbalance interventions and post-encoder corrections.
In each paired comparison, the representation, train/test split, and classifier
protocol are held fixed, and only the LRG residual augmentation is added. This
separates the main question, whether local residuals improve an available
representation, from absolute classifier ranking.

\subsection{Benchmark Construction}
\label{sec:benchmark-construction}
We construct a controlled univariate imbalanced benchmark from the fixed
train/test splits of the UCR archive \cite{dau2019ucr} and 30 univariate
classification datasets released with Bake Off Redux
\cite{middlehurst2024bakeoffredux}. Following prior imbalanced TSC benchmark
construction \cite{zhao2022tsmote}, we keep the original binary or multiclass
label space and downsample only one training class. Let $n_{\max}$ be the
largest class count in the original training split and let $y_{\min}$ denote the
selected minority class. For target ratio $R{:}1$, we keep
$\lfloor n_{\max}/R \rfloor$ training samples from $y_{\min}$ and retain all
other training samples. The test split is left unchanged to isolate minority
training-support scarcity while preserving the fixed evaluation problem.

The paired comparisons use the full $3{:}1$/$5{:}1$/$10{:}1$/$20{:}1$ scope,
giving 138 dataset-ratio tasks per backbone. A dataset-ratio pair is included
only when the original training split can realize the requested ratio, so fewer
datasets are available at higher imbalance. We provide the retained and excluded
dataset-ratio lists on the associated repository.

\begin{table*}[t]
\centering
\footnotesize
\setlength{\tabcolsep}{3.2pt}
\renewcommand{\arraystretch}{1.03}
\caption{Summary statistics for the materialized imbalanced benchmark. Each
column reports the retained dataset-ratio tasks for the target training ratio.
Except for dataset counts, entries are
median (min-max).}
\label{tab:dataset-summary}
\begin{tabular}{lcccc}
\toprule
Statistic & 3:1 & 5:1 & 10:1 & 20:1 \\
\midrule
Retained datasets & 46 & 43 & 29 & 20 \\
Train size & 426 (76-2480) & 420 (90-8898) & 433 (165-8657) & 619 (345-8537) \\
Test size & 383 (61-4000) & 450 (61-8236) & 600 (139-8236) & 780 (184-8236) \\
Classes & 2 (2-42) & 3 (2-8) & 2 (2-7) & 2 (2-7) \\
Series length & 315 (15-4201) & 426 (60-4201) & 256 (60-4201) & 266 (60-4201) \\
Minority train count & 42 (16-620) & 37 (15-481) & 38 (15-240) & 24 (16-120) \\
\bottomrule
\end{tabular}
\end{table*}
\subsection{Representations and Heads}

We evaluate LRG across learned encoders, frozen pretrained features, and
randomised feature-transform representations. Learned encoders include
InceptionTime-style networks \cite{fawaz2020inceptiontime}, ResNet1D, TCN,
TimesNet \cite{wu2023timesnet}, a classification-oriented PatchTST-style
encoder \cite{nie2023patchtst}, and MLP encoders. All learned encoders use the
same inner validation split, early stopping criterion, and plateau scheduling;
weighted variants additionally use class-weighted cross entropy and balanced
sampling where specified. We also evaluate frozen MOMENT features
\cite{goswami2024moment} and features from the combined MultiRocket-Hydra
randomised transform \cite{tan2022multirocket,dempster2023hydra}.

Classifier heads are chosen to match the representation setting. The
learned-encoder, pretrained-feature, and intervention studies use class-weighted
logistic regression. The MultiRocket-Hydra feature-transfer row uses
threshold-tuned logistic regression. The post-encoder correction study fixes the
Inception feature representation and compares calibrated, threshold-tuned, or
retrained classifier heads with and without LRG augmentation.

\subsection{Paired Protocol and Baseline Groups}

For a given representation, train/test split, and classifier head, the paired
protocol compares $\zraw$ with $[\zraw,\rfull(z),\rlda(z)]$. The task, data
split, representation, and classifier protocol are therefore shared; only the
LRG residual augmentation is added. This makes each comparison a direct test of
whether the local residual features improve the available representation.

We use three paired studies. Representation Evaluation
(Table~\ref{tab:lrg-main}) applies the protocol across learned encoders,
pretrained features, and randomised feature-transform representations.
Imbalance-Intervention Compatibility
(Table~\ref{tab:imbalance-interventions}) applies the same Base-versus-+LRG
comparison after losses, samplers, or SMOTE-style over-sampling have changed the
encoder training or training data. Post-Encoder Correction Compatibility
(Table~\ref{tab:post-encoder}) fixes the Inception feature representation and
tests whether residual augmentation remains useful under classifier-level
corrections.

The residual-signal and parameter studies isolate which components carry the
gain, while the analysis section tests whether the gains align with measurable
training-local geometry failure. Balanced accuracy and Macro-F1 are the primary
cross-task metrics; Min-F1, Sens., and Spec. are used for minority-specific
questions.

\section{Experimental Results}

\subsection{Representation Evaluation}

\begin{table*}[t]
\centering
\setlength{\abovecaptionskip}{2pt}
\setlength{\belowcaptionskip}{0pt}
\setlength{\tabcolsep}{4.5pt}
\renewcommand{\arraystretch}{1.0}
\caption{Representation evaluation. W/T/L counts per-task wins, ties, and
losses for +LRG over the corresponding raw
feature. Parentheses in W/T/L cells show Wilcoxon signed-rank evidence levels
($*$: $p<0.05$, $**$: $p<0.01$, $***$: $p<0.001$).}
\label{tab:lrg-main}
\begin{tabular*}{0.95\textwidth}{@{\extracolsep{\fill}}llcccc@{}}
\toprule
Representation & Type & Bal. acc. & W/T/L & Macro-F1 & W/T/L \\
\midrule
Inception & learned & 0.769 $\to$ \textbf{0.786} &
\wtlp{88/4/46}{***} &
0.757 $\to$ \textbf{0.782} & \wtlp{100/3/35}{***} \\
TimesNet & learned & 0.694 $\to$ \textbf{0.737} &
\wtlp{111/0/27}{***} &
0.678 $\to$ \textbf{0.730} & \wtlp{117/0/21}{***} \\
ResNet1D & learned & 0.748 $\to$ \textbf{0.774} &
\wtlp{110/0/28}{***} &
0.733 $\to$ \textbf{0.767} & \wtlp{116/0/22}{***} \\
TCN & learned & 0.710 $\to$ \textbf{0.751} &
\wtlp{109/3/26}{***} &
0.690 $\to$ \textbf{0.742} & \wtlp{116/1/21}{***} \\
PatchTST & learned & 0.742 $\to$ \textbf{0.760} &
\wtlp{86/2/50}{***} &
0.731 $\to$ \textbf{0.754} & \wtlp{92/1/45}{***} \\
MOMENT & frozen pretrained & 0.651 $\to$ \textbf{0.701} &
\wtlp{94/0/44}{***} &
0.619 $\to$ \textbf{0.696} & \wtlp{99/0/39}{***} \\
% \midrule
% MultiRocket-Hydra & native fixed & 0.748 & -- & 0.745 & -- \\
% HC2 & native ensemble & 0.706 & -- & 0.686 & -- \\
MultiRocket-Hydra & randomised transform &
0.765 $\to$ \textbf{0.778} & \wtlp{82/2/54}{*} &
0.749 $\to$ \textbf{0.774} & \wtlp{94/0/44}{***} \\
\bottomrule
\end{tabular*}
\end{table*}

Table~\ref{tab:lrg-main} reports the paired representation evaluation study.
The experiment is not designed as a classifier comparison; it tests whether LRG
improves a given representation under the same split and classifier protocol.
The evidence for LRG is therefore the within-row change from $\zraw$ to $\zlrg$. 

Table~\ref{tab:lrg-main} uses the representation-evaluation training
configuration, where learned encoders are trained with balanced mini-batch
sampling. In contrast, the Nat. rows in
Table~\ref{tab:imbalance-interventions} use ordinary CE with the original
imbalanced dataloader. The two Inception baselines therefore evaluate different
training protocols and are not expected to have identical raw performance.

LRG improves every representation setting in this study. Balanced accuracy increases by 1.3-5.0 percentage points and Macro-F1 by 2.3-7.7 percentage points. The largest gains occur for weaker raw feature spaces: TimesNet and TCN gain more than four points in balanced accuracy and more than five points in Macro-F1, while frozen MOMENT gains five points in balanced accuracy and 7.7 points in Macro-F1. Inception has the strongest raw balanced accuracy and still improves, showing that the effect is not restricted to weak representations. The frozen MOMENT row captures the post-hoc setting most directly: the pretrained encoder supplies fixed features, and LRG augments those features before the classifier is fitted, without updating the encoder. The W/T/L counts show that the gains are broadly distributed across tasks: +LRG wins more tasks than it loses for every learned and frozen row on both metrics, with all such rows significant at $p<0.001$. The randomized-transform row shows the same direction: MultiRocket-Hydra+LRG improves both average metrics, with stronger evidence for Macro-F1 and weaker but positive evidence for balanced accuracy.

\subsection{LRG with Imbalance-Aware Methods}

We next ask whether LRG remains useful after standard imbalance-aware methods have already modified the training data or objective. Table~\ref{tab:imbalance-interventions} pairs each baseline setting with the same setting followed by LRG residual augmentation. The classifier is class-weighted logistic regression for this set of experiments, so the Base-versus-+LRG difference reflects feature augmentation under a fixed decision rule rather than threshold calibration.

\begin{table*}[t]
\centering
\footnotesize
\setlength{\abovecaptionskip}{2pt}
\setlength{\belowcaptionskip}{0pt}
\renewcommand{\arraystretch}{1.05}
\caption{LRG with standard imbalance-aware methods. For this compact
table only, Bal. denotes balanced accuracy, MacF1 denotes Macro-F1, and MinF1
denotes Min-F1.}
\label{tab:imbalance-interventions}
\begin{tabular*}{0.95\textwidth}{@{\extracolsep{\fill}}ll*{12}{c}@{}}
\toprule
\multicolumn{1}{c}{\multirow{2}{*}{Method}} &
\multicolumn{1}{c}{\multirow{2}{*}{Feature}} &
\multicolumn{3}{c}{Inception} &
\multicolumn{3}{c}{ResNet1D} &
\multicolumn{3}{c}{TCN} &
\multicolumn{3}{c}{Avg.} \\
\cmidrule(lr){3-5}\cmidrule(lr){6-8}\cmidrule(lr){9-11}\cmidrule(lr){12-14}
 & & Bal. & MacF1 & MinF1
 & Bal. & MacF1 & MinF1
 & Bal. & MacF1 & MinF1
 & Bal. & MacF1 & MinF1 \\
\midrule
Nat. & Base
& 0.762 & 0.749 & 0.713
& 0.740 & 0.724 & 0.674
& 0.701 & 0.680 & 0.633
& 0.734 & 0.718 & 0.674 \\
 & +LRG
& \textbf{0.790} & \textbf{0.784} & \textbf{0.745}
& \textbf{0.770} & \textbf{0.762} & \textbf{0.716}
& \textbf{0.752} & \textbf{0.743} & \textbf{0.702}
& \textbf{0.771} & \textbf{0.763} & \textbf{0.721} \\
BS & Base
& 0.768 & 0.757 & 0.718
& 0.748 & 0.732 & 0.687
& 0.704 & 0.685 & 0.646
& 0.740 & 0.725 & 0.684 \\
 & +LRG
& \textbf{0.789} & \textbf{0.783} & \textbf{0.742}
& \textbf{0.773} & \textbf{0.765} & \textbf{0.720}
& \textbf{0.754} & \textbf{0.746} & \textbf{0.710}
& \textbf{0.772} & \textbf{0.765} & \textbf{0.724} \\
LDAM & Base
& 0.766 & 0.755 & 0.721
& 0.757 & 0.744 & 0.700
& 0.708 & 0.689 & 0.654
& 0.744 & 0.729 & 0.692 \\
 & +LRG
& \textbf{0.788} & \textbf{0.782} & \textbf{0.747}
& \textbf{0.776} & \textbf{0.769} & \textbf{0.725}
& \textbf{0.755} & \textbf{0.746} & \textbf{0.714}
& \textbf{0.773} & \textbf{0.766} & \textbf{0.729} \\
SMOTE & Base
& 0.765 & 0.752 & 0.716
& 0.745 & 0.728 & 0.689
& 0.705 & 0.685 & 0.649
& 0.739 & 0.722 & 0.685 \\
 & +LRG
& \textbf{0.785} & \textbf{0.779} & \textbf{0.739}
& \textbf{0.770} & \textbf{0.762} & \textbf{0.720}
& \textbf{0.748} & \textbf{0.739} & \textbf{0.704}
& \textbf{0.768} & \textbf{0.760} & \textbf{0.721} \\
e-SMOTE & Base
& 0.764 & 0.750 & 0.718
& 0.739 & 0.721 & 0.686
& 0.705 & 0.684 & 0.655
& 0.736 & 0.718 & 0.687 \\
 & +LRG
& \textbf{0.784} & \textbf{0.777} & \textbf{0.741}
& \textbf{0.765} & \textbf{0.757} & \textbf{0.719}
& \textbf{0.748} & \textbf{0.739} & \textbf{0.710}
& \textbf{0.766} & \textbf{0.758} & \textbf{0.723} \\
\bottomrule
\end{tabular*}
\vspace{-0.25cm}
\end{table*}
Rows are paired by baseline setting, so the LRG delta is the within-pair
difference; bold marks the better value in each pair. The Avg. block reports
the mean across the three backbone columns. BS denotes Balanced Softmax
\cite{ren2020balancedsoftmax}, LDAM denotes LDAM-DRW
\cite{cao2019ldam}, and Nat. denotes ordinary CE. Nat., BS, and LDAM are
trained on the original imbalanced dataloader; SMOTE and e-SMOTE are
input-space over-sampling methods paired with weighted CE.

LRG improves all baseline settings and backbones. In the Avg. block, adding LRG improves balanced accuracy by 2.9-3.7 percentage points, Macro-F1 by 3.7-4.5 points, and Min-F1 by 3.6-4.7 points. The gains are largest for TCN, where the raw representation is less stable: across imbalance-aware methods, TCN gains 4.3-5.1 points in balanced accuracy, 5.4-6.3 points in Macro-F1, and 5.5-6.9 points in Min-F1. The improvements remain visible for ResNet1D and Inception, including after Balanced Softmax, LDAM, SMOTE, and e-SMOTE. These results suggest that imbalance-aware losses and over-sampling can improve
the learned representation while still leaving useful local residual information
for LRG to exploit.

\begin{figure*}[t]
\centering
\begin{minipage}{0.32\textwidth}
\centering
\includegraphics[width=\linewidth]{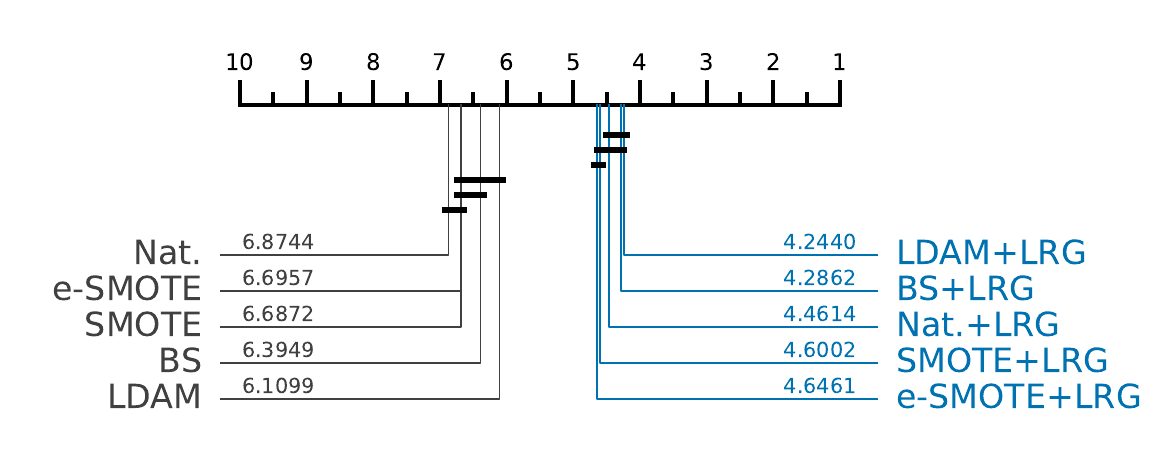}\\[-0.5ex]
{\scriptsize (a) Balanced accuracy}
\end{minipage}\hfill
\begin{minipage}{0.32\textwidth}
\centering
\includegraphics[width=\linewidth]{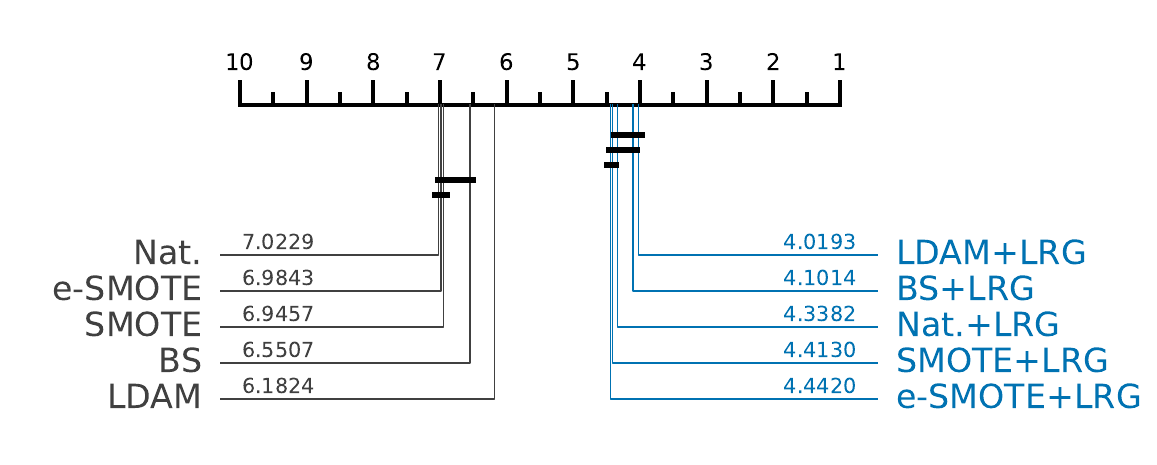}\\[-0.5ex]
{\scriptsize (b) Macro-F1}
\end{minipage}\hfill
\begin{minipage}{0.32\textwidth}
\centering
\includegraphics[width=\linewidth]{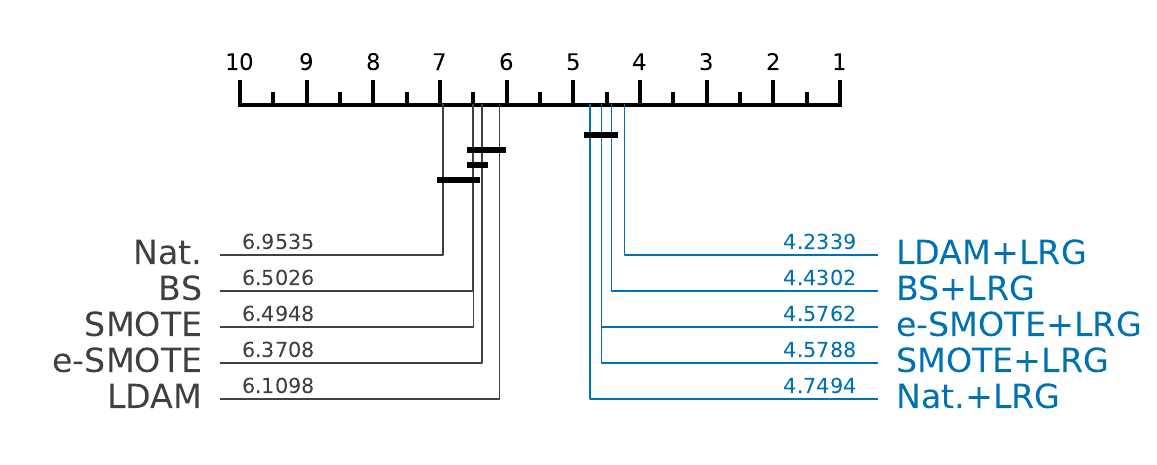}\\[-0.5ex]
{\scriptsize (c) Min-F1}
\end{minipage}
\caption{Average rank diagrams for the intervention rows in
Table~\ref{tab:imbalance-interventions}.}
\label{fig:cd-interventions}
\vspace{-0.5cm}
\end{figure*}

Figure~\ref{fig:cd-interventions} ranks the methods over the same 138 backbone--dataset--ratio problems as Table~\ref{tab:imbalance-interventions}. Lower rank is better. Cliques indicate methods that are not separated by a one-sided Wilcoxon signed-rank test with Holm correction at $\alpha=0.1$. The leading group consists of +LRG variants; small rank reversals between BS+LRG and LDAM+LRG should be read as metric-specific differences rather than as a stable ordering between the two losses.

\subsection{Post-Encoder Classifier Corrections}

We next fix the Inception feature representation and ask whether LRG remains useful when the classifier has already been corrected. Thresholds, calibration parameters, and
predicted-prior estimates are selected using training-only out-of-fold (OOF)
predictions, with no access to test labels or the test distribution. For +LRG, all LRG
components are fit within each OOF training fold. Unlike
Table~\ref{tab:imbalance-interventions}, which changes the training data or
objective, Table~\ref{tab:post-encoder} changes only the downstream decision
rule. The key comparison is the paired Base-versus-+LRG difference within each
correction.

\begin{table*}[t]
\centering
\footnotesize
\setlength{\abovecaptionskip}{2pt}
\setlength{\belowcaptionskip}{0pt}
\setlength{\tabcolsep}{4.5pt}
\renewcommand{\arraystretch}{1.05}
\caption{Compatibility with post-encoder classifier corrections on Inception
features.}
\label{tab:post-encoder}
\begin{tabular*}{0.95\textwidth}{@{\extracolsep{\fill}}llcccccc@{}}
\toprule
Method & Level & Bal. acc. & $\Delta$ bal. (\%) &
Macro-F1 & $\Delta$ Macro-F1 (\%) &
Min-F1 & $\Delta$ Min-F1 (\%) \\
\midrule
% Unweighted & imbalance-naive head & 0.692$\to$\textbf{0.755} & +6.29 &
% 0.662$\to$\textbf{0.752} & +8.94 &
% 0.491$\to$\textbf{0.657} & +16.64 \\
% \midrule
CW-logistic & class-weighted head & 0.762$\to$\textbf{0.790} & +2.80 &
0.749$\to$\textbf{0.784} & +3.50 &
0.713$\to$\textbf{0.745} & +3.20 \\
Threshold & threshold & 0.764$\to$\textbf{0.785} & +2.00 &
0.757$\to$\textbf{0.781} & +2.47 &
0.714$\to$\textbf{0.738} & +2.36 \\
LogitAdj & logit prior & 0.766$\to$\textbf{0.788} & +2.17 &
0.757$\to$\textbf{0.782} & +2.50 &
0.715$\to$\textbf{0.738} & +2.30 \\
P2P & pred. prior & 0.765$\to$\textbf{0.788} & +2.22 &
0.756$\to$\textbf{0.782} & +2.62 &
0.716$\to$\textbf{0.738} & +2.25 \\
VectorScale & calibration & 0.768$\to$\textbf{0.790} & +2.16 &
0.758$\to$\textbf{0.786} & +2.77 &
0.722$\to$\textbf{0.743} & +2.15 \\
TauNorm & weight norm. & 0.688$\to$\textbf{0.752} & +6.36 &
0.657$\to$\textbf{0.748} & +9.04 &
0.479$\to$\textbf{0.647} & +16.82 \\
cRT & classifier retrain & 0.770$\to$\textbf{0.789} & +1.82 &
0.759$\to$\textbf{0.785} & +2.59 &
0.725$\to$\textbf{0.743} & +1.85 \\
\bottomrule
\end{tabular*}
\vspace{-0.5cm}
% \vspace{0.5mm}

% \begin{minipage}{0.75\textwidth}
% \scriptsize
% \emph{Note.} Unweighted is included as an imbalance-naive reference. The
% CW-logistic row is the Nat./Inception pair from
% Table~\ref{tab:imbalance-interventions}.
% \end{minipage}
\end{table*}

Base denotes each correction fitted on raw Inception features, while +LRG fits
the same correction type on LRG-augmented features under the same training-only
protocol. The Level column identifies which part of the frozen-feature decision
pipeline is changed. Threshold tuning selects a minority threshold from
out-of-fold probabilities. LogitAdj applies training-prior logit adjustment
\cite{menon2021logitadjustment}; P2P replaces the empirical prior with the
out-of-fold predicted prior of the model, following the Prior2Posterior
correction idea \cite{bhat2025prior2posterior}. VectorScale is a regularized
class-specific affine scaling of logits inspired by post-hoc calibration
methods \cite{guo2017calibration}. TauNorm rescales classifier weights, and
cRT retrains only the classifier with class-balanced weighting, following
decoupled classifier correction ideas \cite{kang2020decoupling}.

LRG improves balanced accuracy, Macro-F1, and Min-F1 for every post-encoder
correction in this table. The largest deltas occur for TauNorm, whose raw row is
the weakest in the table, but the more informative result is that LRG also
improves stronger classifier-level corrections: VectorScale gains 2.16 points
in balanced accuracy, 2.77 points in Macro-F1, and 2.15 points in Min-F1, while
cRT gains 1.82, 2.59, and 1.85 points, respectively. Thus, classifier-level
correction and residual augmentation are complementary: the former changes the
decision rule on fixed features, whereas LRG exposes additional local residual
coordinates before the head is fit. 

\subsection{Residual Signal and Sensitivity}
This subsection asks whether LRG gains arise specifically from signed local
residuals rather than from generic local descriptors, and then examines the
contribution and parameter sensitivity of the residual coordinates. All tables
use the same 138 dataset--ratio tasks and the same cached Inception CE
representations. Reported values are paired percentage-point changes relative
to raw features under the same downstream head. For each configuration, we
first average the class-weighted logistic and threshold-tuned deltas within
each task and then average across tasks.
Table~\ref{tab:local-feature-controls} compares LRG with alternative local
features, while Table~\ref{tab:residual-sensitivity} reports residual ablations
and one-factor-at-a-time parameter sensitivity.

\begin{table}[t]
\centering
\scriptsize
\setlength{\tabcolsep}{2.0pt}
\renewcommand{\arraystretch}{1.04}
\caption{Local-feature controls.}
\label{tab:local-feature-controls}
\begin{tabular}{lccccc}
\toprule
Added signal & $\Delta$ Bal. & $\Delta$ Macro-F1 & $\Delta$ Min-F1 & $\Delta$ Sens. & $\Delta$ Spec. \\
\midrule
Class-proto dist. & +0.14 & +0.40 & +0.73 & -0.26 & +0.62 \\
$k$-means dist. & +0.35 & +0.63 & +0.97 & -0.10 & +0.83 \\
RBF affinity & +0.06 & +0.31 & +0.45 & -0.65 & +0.68 \\
Scalar stats & +0.10 & +0.40 & +0.83 & -0.37 & +0.71 \\
RBF + stats & +0.01 & +0.30 & +0.59 & -0.74 & +0.78 \\
VLAD & -8.83 & -8.17 & -6.49 & -13.99 & \textbf{+4.16} \\
LRG residuals (ours) & \textbf{+2.27} & \textbf{+3.07} & \textbf{+4.02} & \textbf{+1.05} & +3.16 \\
\bottomrule
\end{tabular}
\vspace{-0.25cm}
\end{table}

\begin{table}[t]
\centering
\scriptsize
\setlength{\tabcolsep}{3.0pt}
\renewcommand{\arraystretch}{1.02}
\caption{Residual ablation and parameter sensitivity.}
\label{tab:residual-sensitivity}
\begin{tabular}{lccc}
\toprule
% \multicolumn{4}{l}{\emph{Residual ablation}} \\
Configuration & $\Delta$ Bal. & $\Delta$ Macro-F1 & $\Delta$ Min-F1 \\
\midrule
$\rfull$ & +2.11 & +2.87 & +3.88 \\
$\rlda$ & +1.88 & +2.59 & +3.45 \\
$\rfull+\rlda$ & \textbf{+2.27} & \textbf{+3.07} & \textbf{+4.02} \\
\midrule
\multicolumn{4}{l}{\emph{Number of local references}} \\
$K=4$ & +2.25 & +3.02 & +3.99 \\
$K=8$ (default) & +2.27 & +3.07 & +4.02 \\
$K=16$ & +2.33 & +3.12 & +4.12 \\
\addlinespace[1pt]
\multicolumn{4}{l}{\emph{LDA residual dimension}} \\
$d_{\mathrm{LDA}}\le 1$ (default) & +2.27 & +3.07 & +4.02 \\
$d_{\mathrm{LDA}}\le 2$ & +2.35 & +3.16 & +4.19 \\
$d_{\mathrm{LDA}}\le 3$ & +2.41 & +3.21 & +4.27 \\
\addlinespace[1pt]
\multicolumn{4}{l}{\emph{Affinity temperature}} \\
$\tau=0.5$ & +1.85 & +2.64 & +3.44 \\
$\tau=1.0$ (default) & +2.27 & +3.07 & +4.02 \\
$\tau=2.0$ & +2.10 & +2.88 & +3.83 \\
\bottomrule
\end{tabular}
\vspace{-0.5cm}
\end{table}

Table~\ref{tab:local-feature-controls} treats the complete LRG residual
augmentation as the proposed signal and compares it against local-feature
alternatives. Prototype distances, $k$-means distances, RBF-style affinities, and
scalar local statistics give small gains, while the VLAD-style replacement code
improves specificity at the cost of sensitivity and the main metrics. The LRG
row improves all main metrics and sensitivity while preserving the original
feature. This supports the directionality principle in
Section~\ref{sec:geometric-effect}: local distances and reliability statistics
can mark a region as nearby, mixed, or sparse, but they do not expose the signed
direction of the sample's deviation from the local training frame.

The residual ablation block in Table~\ref{tab:residual-sensitivity} compares
the two residual components and their combination. Both $\rfull$ and $\rlda$
are useful on their own, and appending them together gives the strongest
balanced accuracy, Macro-F1, and Min-F1.

The sensitivity blocks are one-factor-at-a-time robustness checks rather than
hyperparameter tuning procedures. The main experiments use the fixed defaults
$K=8$, $d_{\mathrm{LDA}}=1$, and $\tau=1.0$. All tested settings retain
positive aggregate gains. Performance varies only modestly across
$K\in\{4,8,16\}$, while $\tau=1.0$ gives the strongest result among the tested
temperatures. Increasing $d_{\mathrm{LDA}}$ from one to three produces small
additional gains on the multiclass tasks. For binary tasks, all three settings
reduce to the single available discriminant direction because
$d_{\mathrm{LDA}}\le C-1$. We retain $d_{\mathrm{LDA}}=1$ as a compact,
pre-specified default that avoids dataset-specific dimensionality selection;
we do not claim that it is the empirically optimal value.

Finally, a descriptive margin decomposition shows that the fitted full-LRG
heads actively use the residual coordinates. Under standardized features, the
$\zraw/\rfull/\rlda$ blocks account for $3.9\%/60.9\%/35.2\%$ of the absolute
minority-versus-strongest-rest margin for class-weighted logistic regression,
and $3.6\%/61.7\%/34.8\%$ for threshold tuning. Because these blocks are
correlated and $\rlda$ is derived from $\rfull$, the percentages are not unique
causal attributions. The following analysis examines where this residual use
improves predictions.

\section{Analysis}

The paired experiments above establish the performance effect of LRG. We next
ask whether this effect is tied to the proposed training-local geometry failure.
The analysis starts with the feature space itself: does imbalance make minority
neighborhoods more rest-exposed? It then tests whether training-only
class-composition risk identifies error-prone regions and whether LRG gains are
larger in those regions. We close by examining how downstream heads trade
sensitivity against specificity when imbalance becomes severe.

\subsection{RQ1: Does Imbalance Alter Local Geometry?}

Before applying any correction, we test whether stronger imbalance changes the
local geometry around the training-defined minority class. We compute RRE
according to Eq.~\ref{eq:rre} before fitting any downstream head. The
neighborhoods are computed after the same training-split standardization and
$\ell_2$ normalization used by LRG, so they should be read as an operational
local-support diagnostic rather than as a claim that a fixed-transform
classifier itself is nearest-neighbor based.

Table~\ref{tab:rq1} compares RRE across imbalance ratios and feature families.
The fixed row summarizes Hydra and MultiRocket fixed-transform features, the
learned row summarizes Inception and PatchTST features trained with weighted CE,
and the pretrained row uses frozen MOMENT features. RRE increases in all three
feature families as imbalance becomes more severe. Because RRE divides the
local rest-neighbor fraction by the split-specific rest-class prior, this trend
is not the mechanical effect of having more rest-class samples in the
training split. The fixed and pretrained rows are especially useful controls:
their extractors are unchanged as the imbalance ratio varies, so the increase
reflects support thinning in the current imbalanced training set. As minority
training samples are removed, minority neighborhoods lose same-class support and
move toward the split-specific random-prior baseline, even though the absolute
RRE values remain below one. The learned row has lower absolute RRE, as expected
for task-supervised features trained with labels to be more class-aligned under
this local metric; nevertheless, it shows the same support-local effect in a
feature space also shaped by imbalance-aware supervised training.
This motivates the next question: whether these entangled local regions
correspond to actual prediction errors, and whether LRG residuals are useful
precisely where the local training geometry is unreliable.

\begin{table}[t]
\centering
\footnotesize
\renewcommand{\arraystretch}{1.05}
\caption{Relative rest exposure (RRE) across feature families.}
\label{tab:rq1}
\begin{tabular}{lcccc}
\toprule
Family & 3:1 & 5:1 & 10:1 & 20:1 \\
\midrule
Fixed & 0.399 & 0.414 & 0.484 & 0.590 \\
Learned & 0.192 & 0.258 & 0.350 & 0.421 \\
Pretrained & 0.578 & 0.656 & 0.694 & 0.741 \\
\bottomrule
\end{tabular}
\vspace{-0.5cm}
\end{table}
\subsection{RQ2: Do Local Risks Track Errors and LRG Gains?}

RQ1 shows that stronger imbalance makes minority neighborhoods more
rest-exposed. We now ask whether the same training-local frame can identify
which regions are error-prone and whether LRG helps most in those regions.
These risk scores use only class-composition statistics from training features,
before test labels are observed.
Within each local reference region, we score class mixture using the local
minority fraction, rest fraction, minority-versus-rest ambiguity, and multiclass
impurity. The false-positive risk is high for rest samples near ambiguous
minority support; the false-negative risk is high for minority samples near
ambiguous rest support; and the hybrid risk combines these two directions with
region impurity. A test feature inherits these region scores through the same
training-derived affinities used by the residual. These scalar diagnostics share
LRG's reference frame but use class composition rather than signed displacement,
so enrichment is not guaranteed by construction. Test labels are used only
afterward to check whether high-risk regions indeed contain more errors.
Using the same feature-family grouping as Table~\ref{tab:rq1},
Table~\ref{tab:rq2} reports three summaries. FP enrichment is the high-to-low
false-positive ratio after binning by false-positive risk. The low/high error
columns report total error rates after binning by hybrid risk, and positive FP
enrichment excludes tasks whose low-risk bin has zero false positives.

\begin{table}[t]
\centering
\footnotesize
\renewcommand{\arraystretch}{1.05}
\caption{High-risk local regions enrich errors under threshold tuning. Error
rates are percentages.}
\label{tab:rq2}
\begin{tabular}{@{}lcccc@{}}
\toprule
Family & FP enr. & Pos.-FP enr. & Low err. & High err. \\
\midrule
Fixed & 1.87$\times$ & 4.53$\times$ & 19.69 & 22.53 \\
Learned & 1.15$\times$ & 6.07$\times$ & 22.14 & 29.03 \\
Pretrained & 3.57$\times$ & 3.55$\times$ & 24.47 & 29.36 \\
\bottomrule
\end{tabular}
\vspace{-0.25cm}
\end{table}

Table~\ref{tab:rq2} shows that high-risk regions are not arbitrary partitions.
High FP-risk regions enrich false positives, especially when conditioning on
tasks where false positives exist in the low-risk reference bin. High
hybrid-risk regions also have higher total error rates in every feature family.
This supports the diagnostic claim: training-local geometry can stratify test
regions by error risk before test labels are used for evaluation.

We then test whether the intervention is aligned with this diagnostic. For each
test sample, we compute its training-only hybrid risk score, split samples into
low- and high-risk bins within each dataset-ratio task, and compare raw and LRG
predictions within each bin. Table~\ref{tab:risk-gain} shows that the high-risk
bin has both higher raw error and a larger paired LRG error reduction. The gain
is therefore not only an average effect over all samples; it is amplified where
the local training geometry is more entangled. The bin-level
decomposition shows that this concentration is especially strong for
false-positive repair, motivating the head-level safety analysis in RQ3.

\begin{table}[t]
\centering
\footnotesize
\renewcommand{\arraystretch}{1.05}
\caption{Risk-bin gain analysis with class-weighted logistic.}
\label{tab:risk-gain}
\begin{tabular}{llcccc}
\toprule
Backbone & Bin & Raw err. (\%) & LRG err. (\%) & $\Delta$ err. & $\Delta$ FP \\
\midrule
Inception & Low & 19.30 & 17.32 & +1.97 & +1.13 \\
Inception & High & 28.63 & 24.09 & +4.55 & +6.98 \\
ResNet1D & Low & 21.92 & 19.37 & +2.55 & +1.63 \\
ResNet1D & High & 31.65 & 26.03 & +5.62 & +6.88 \\
TCN & Low & 24.71 & 21.12 & +3.58 & +1.84 \\
TCN & High & 34.67 & 27.20 & +7.47 & +7.94 \\
\bottomrule
\end{tabular}
\vspace{-0.5cm}
\end{table}

\begin{figure}[htbp]
\centering
\includegraphics[width=\columnwidth]{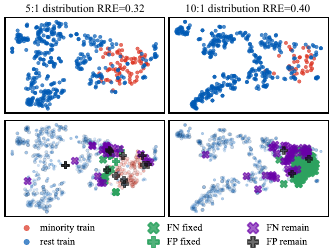}
\caption{LRG error repair on Strawberry at 5:1 and 10:1. The upper row shows
minority/rest features; the lower row overlays error transitions on the same
t-SNE coordinates.}
\label{fig:lrg-tsne-grid}
\end{figure}

As a qualitative check, Figure~\ref{fig:lrg-tsne-grid} visualizes the same
Inception CE feature map on Strawberry under 5:1 and 10:1 imbalance. The upper
row shows the minority/rest structure in the raw t-SNE coordinates. The lower
row overlays LRG error transitions on the same coordinates: green markers are
false negatives or false positives corrected by LRG, while purple and black
markers are false negatives and false positives that remain after LRG. The
figure does not require global separation in the plotted embedding. Instead, it
shows whether errors in locally mixed regions are corrected on the same map. At
5:1, LRG makes a smaller but consistent change, reducing false negatives from
25 to 21 and false positives from 14 to 12. At 10:1, where RRE is higher
($0.40$ versus $0.32$), the green false-negative repairs are more prominent:
false negatives decrease from 132 to 40, with false positives increasing from
0 to 3. This visualizes the risk-bin result that LRG gains concentrate in
locally mixed minority regions.

\subsection{RQ3: High-Imbalance Behavior and Head-Level Safety}

To isolate ratio effects from changes in which source datasets are
averaged, this subsection uses the matched subset of $14$ datasets that appear
under all four ratios, 3:1, 5:1, 10:1, and 20:1. We summarize LRG behavior by
averaging paired deltas over five post-encoder settings: linear, threshold
tuning, logit adjustment, vector scaling, and cRT. In
Table~\ref{tab:safety-tradeoff}, RRE is computed on the corresponding Inception
CE diagnostic run; the remaining columns are paired mean deltas in percentage
points.
\begin{table}[t]
\centering
\footnotesize
\caption{Ratio-level LRG effect on the matched $14$-dataset scope.}
\label{tab:safety-tradeoff}
\begin{tabular}{cccccc}
\toprule
Ratio & RRE & $\Delta$ Bal. & $\Delta$ Min-F1 & $\Delta$ Sens. & $\Delta$ Spec. \\
\midrule
3:1  & 0.299 & +0.87 & +1.25 & -0.08 & +1.80 \\
5:1  & 0.323 & +0.80 & +1.46 & -0.02 & +1.53 \\
10:1 & 0.342 & +2.35 & +4.08 & +1.85 & +2.86 \\
20:1 & 0.454 & +2.87 & +6.65 & +2.64 & +2.91 \\
\bottomrule
\end{tabular}
\vspace{-0.5cm}
\end{table}

On this matched $14$ dataset scope, Table~\ref{tab:safety-tradeoff} shows a
clear high-imbalance trend. RRE rises
from $0.299$ at 3:1 to $0.454$ at 20:1, showing that minority neighborhoods
become increasingly rest-exposed even after normalizing by the global rest-class
prior. LRG gains increase in the same high-imbalance regimes. At 20:1, LRG
improves balanced accuracy by $+2.87$pp, Min-F1 by $+6.65$pp, sensitivity by
$+2.64$pp, and specificity by $+2.91$pp on average across the five post-encoder
settings.
Thus, the high-imbalance advantage is not merely a specificity-driven
false-positive repair: sensitivity and specificity both improve. This
is consistent with the geometric interpretation. As imbalance removes minority
support, raw local neighborhoods become harder to separate from $\zraw$ alone;
the signed training-referenced residual then provides a larger corrective signal
for the downstream head. The near-zero sensitivity deltas at 3:1 and 5:1 are
also informative: when the base post-encoder settings already preserve minority
recall reasonably well, LRG improves the local boundary without materially
changing the minority-recall operating point.

\section{Discussion}

Taken together, the results support a focused claim: imbalance induces a
measurable training-local geometry failure in otherwise useful time series
representations, and signed local residuals expose corrective information for a
lightweight downstream head. LRG does not treat the raw representation as failed
globally. Instead, it asks whether minority support is locally unreliable after a
feature space is already available, then supplies a directional coordinate tied
to nearby training geometry. This post-hoc setting is practical when pretrained
features are expensive to fine-tune or only features are available.

This distinction also separates diagnostics from repair. Reliability, affinity, and risk
statistics summarize whether a region is pure, mixed, sparse, or error-prone,
but they do not specify a discriminative direction. A local residual is a signed
coordinate: it tells the head how a sample is displaced relative to nearby
training geometry. In this sense, diagnostics locate the failure, while the
residual provides the repair coordinate.

LRG should therefore be viewed as a representation-level intervention rather
than another sampler, loss, or threshold rule. Over-sampling changes the training
distribution, class weighting changes the loss, and threshold tuning changes the
final decision boundary. LRG instead exposes a local residual signal that may
remain after those mechanisms, while the downstream head remains responsible for
the sensitivity--specificity operating point. This explains why LRG is
complementary to long-tailed losses, over-sampling, and post-encoder
corrections rather than a replacement for them. The added post-encoder cost is
modest in our feature-space experiments: excluding encoder inference, average
LRG fitting takes 0.433s on Inception features and 1.676s on MOMENT features,
with microsecond-scale prediction overhead.

Using $k$-means centers and a scalar radial scale keeps the post-hoc state small
and avoids unstable covariance estimates under sparse minority support, but it
also limits the ability to model anisotropic local geometry. When minority
support is extremely sparse or highly diffuse, rest-dominated anchors may yield
unstable residual directions, and the LDA projection can become high-variance.
The present benchmark is also univariate by
design: the fixed UCR and Bake Off Redux splits provide a broad controlled
setting for materializing label imbalance, but multivariate archives remain an
important next step. Because LRG operates after feature extraction, the main
question there is how channel interactions affect the learned local reference
geometry. These limitations point to adaptive reference construction and
training-time objectives for minority-reliable local geometry as natural future
directions.

\section{Conclusion}

We introduced Local Reference Geometry as a diagnostic-and-repair framework for
imbalanced time series representations. LRG quantifies training-local geometry
failure through exposure and class-mixture risk, then appends signed full and
LDA-projected residual coordinates to the existing feature. Experiments show
that this directional residual signal provides useful corrective information
across learned, pretrained, and fixed representations, and that the largest
gains occur in local regions identified as risky by the same training-only
frame. Future work can extend this local-reliability view to training-time
objectives for minority-reliable representation learning.

\bibliographystyle{IEEEtran}
\bibliography{references}

\end{document}